\documentclass[10pt,twocolumn,letterpaper]{article}

\usepackage{iccv}
\usepackage{times}
\usepackage{epsfig}
\usepackage{graphicx}
\usepackage{amsmath}
\usepackage{amssymb}
\usepackage{color}
\usepackage{pifont}
\usepackage{multirow}
\usepackage{comment}
\usepackage{threeparttable}
\newcommand{\tabincell}[2]{\begin{tabular}{@{}#1@{}}#2\end{tabular}}

\usepackage[pagebackref=true,breaklinks=true,letterpaper=true,colorlinks,bookmarks=false]{hyperref}

\iccvfinalcopy 

\def\iccvPaperID{6676} 

\ificcvfinal\fi

\begin{document}
	
	\title{Learning A Single Network for Scale-Arbitrary Super-Resolution}
	
	\author{
		Longguang Wang, 
		Yingqian Wang, 
		Zaiping Lin, 
		Jungang Yang, 
		Wei An, 
		Yulan Guo$^{*}$\\
		National University of Defense Technology\\
		{\tt\small \{wanglongguang15,yulan.guo\}@nudt.edu.cn}}
	
	\maketitle
	\ificcvfinal\thispagestyle{empty}\fi
	
	\begin{abstract}
		Recently, the performance of single image super-resolution (SR) has been significantly improved with powerful networks. However, these networks are developed for image SR with specific integer scale factors ({e.g.}, $\times2/3/4$), and cannot handle non-integer and asymmetric SR.
		In this paper, we propose to learn a scale-arbitrary image SR network from scale-specific networks. Specifically, we develop a plug-in module
		for existing SR networks to perform scale-arbitrary SR, which consists of multiple scale-aware feature adaption blocks and a scale-aware upsampling layer. 
		Moreover, conditional convolution is used in our plug-in module to generate dynamic scale-aware filters, which enables our network to adapt to  arbitrary scale factors. 
		Our plug-in module can be easily adapted to existing networks to realize scale-arbitrary SR with a single model. 
		These networks plugged with our module can produce promising results for non-integer and asymmetric SR while maintaining state-of-the-art performance for SR with integer scale factors. Besides, the additional computational and memory cost of our module is very small.
		
	\end{abstract}
	
	\section{Introduction}
	Single image super-resolution (SR) aims at recovering a high-resolution (HR) image from its low-resolution (LR) counterpart. As a long-standing low-level computer vision problem, single image SR has been investigated for decades \cite{Sun2008Image,Zhang2012Single,Yang2013Fast,Timofte2013Anchored,Gu2015Convolutional,Haris2018Deep}. Recently, the rise of deep learning provides a powerful tool to solve this problem, with numerous CNN-based methods \cite{Dong2014Learning,Kim2016Accurate,Caballero2017Real,Zhang2018Residual,Qiu2019Embedded} being developed to improve the SR performance.
	
	Although recent CNN-based single image SR networks \cite{Haris2018Deep,Zhang2018Residual,Zhang2018Image,Dai2019Second} have achieved promising performance, they are developed for image SR with specific integer scale factors (\emph{e.g.}, $\times2/3/4$). 
	\textcolor{black}{In many real-world applications like image retargeting, image editing and artworks, non-integer SR (\emph{e.g.}, from $100\times100$ to $220\times220$) and asymmetric SR (\emph{e.g.}, from $100\times100$ to $220\times420$) are highly demanded.}
	However, due to the fixed filters in upscale modules, most existing networks can only zoom in an image with specific integer scales and cannot handle scale-arbitrary SR in real-world scenarios. 
	
	\begin{figure}[t]
		\centering
		\includegraphics[width=1\linewidth]{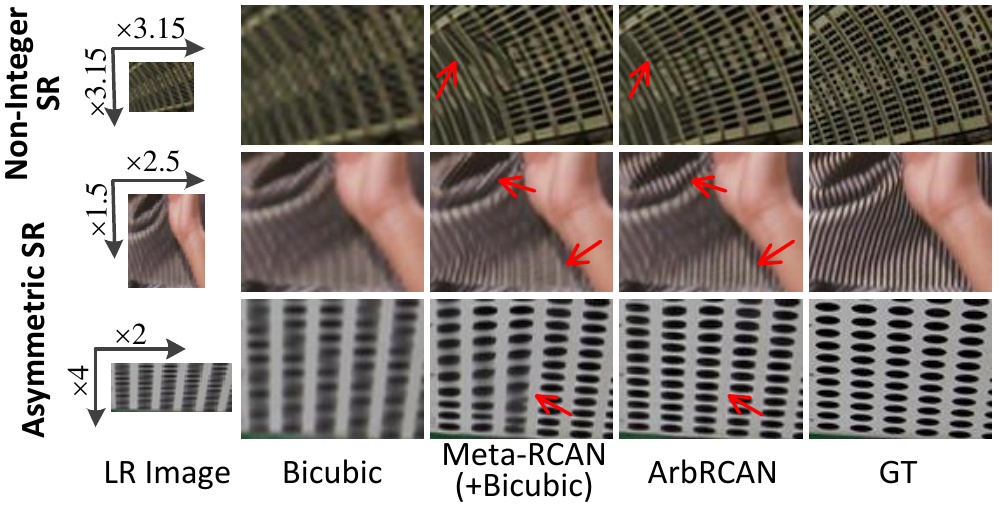}
		\caption{Visual comparison achieved by Bicubic, Meta-RCAN \cite{Hu2019Meta} and our ArbRCAN. ``(+Bicubic)'' means that the network output is further resized to the expected resolution using bicubic interpolation for asymmetric scale factors.}
		\label{fig0}
		\vspace{-0.2cm}
	\end{figure}
	
	To address this limitation, Hu \emph{et al.} \cite{Hu2019Meta} proposed a Meta-SR network to dynamically predict filters in the upscale module for different scale factors using meta-learning. As a result, Meta-SR achieves promising performance on non-integer scale factors. However, scale information is only used for upsampling in Meta-SR. That is, features in the backbone are the same for all SR tasks with different scale factors, which hinders the further improvement of performance. Moreover, Meta-SR  focuses on SR with non-integer scale factors but cannot handle SR with asymmetric scale factors.
	
	In this paper, we propose to learn a scale-arbitrary single image SR network from scale-specific networks. Specifically, we develop a plug-in module for existing SR networks to enable scale-arbitrary SR, which consists of multiple scale-aware feature adaption blocks and a scale-aware upsampling layer. The scale-aware feature adaption blocks are used to adapt features in the backbone to specific scale factors and the scale-aware upsampling layer is used for scale-arbitrary upsampling. 
	Within our plug-in module, conditional convolutions are used to generate dynamic scale-aware filters to handle different scale factors.
	Our plug-in module can be easily adapted to existing networks for scale-arbitrary SR with small additional computational and memory cost. Baseline networks equipped with our module can produce promising results for non-integer and asymmetric SR (Fig.~\ref{fig0}), while maintaining state-of-the-art performance on integer scale factors with a single model. To the best of our knowledge, our plug-in module is the first work to handle asymmetric SR.
	
	Our main contributions can be summarized as follows: 
	1) We develop a plug-in module for existing SR networks to achieve scale-arbitrary SR, including multiple scale-aware feature adaption blocks and a scale-aware upsampling layer. 
	2) Our plug-in module uses conditional convolution to dynamically generate filters  based on the input scale information, which facilitates our network to adapt to specific scale factors. 
	3) Experimental results show that baseline networks equipped with our module produce promising results for scale-arbitrary SR with only a single model. A video demo is provided in the supplemental material.
	
	\section{Related Work}
	In this section, we first briefly review several major works for CNN-based single image SR. Then, we discuss conditional convolutions that are related to our work.
	
	\noindent \textbf{Single Image Super-Resolution.}
	Due to the powerful feature representation and model fitting capabilities of deep neural network, CNN-based single image SR methods \cite{Dong2014Learning,Kim2016Accurate,Caballero2017Real,Haris2018Deep,Zhang2018Residual} outperform traditional methods \cite{Sun2008Image,Zhang2012Single,Yang2012Coupled,Yang2013Fast,Timofte2013Anchored,Gu2015Convolutional} significantly. Dong \emph{et al.} \cite{Dong2014Learning} proposed a three-layer convolutional network (namely, SRCNN) to learn the non-linear mapping between LR images and HR images. A deeper network (namely, VDSR) with 20 layers   \cite{Kim2016Accurate} was then developed to achieve better performance. 
	Later, Lim \emph{et al.} \cite{Lim2017Enhanced} proposed a very deep and wide network, namely EDSR.
	Specifically, batch normalization (BN) layers were removed and a residual scaling technique was used to enable the training of such a large model. 
	Recently, Zhang \emph{et al.} \cite{Zhang2018Image} and Dai \emph{et al.} \cite{Dai2019Second} further improved the SR performance by introducing channel attention and second-order channel attention, respectively.
	
	Although existing single image SR networks have achieved promising results, they are trained for SR with a single specific integer scale factor. To overcome this limitation, Lim \emph{et al.} \cite{Lim2017Enhanced} proposed a multi-scale deep super-resolution (MDSR) system  to integrate modules trained for multiple integer scale factors (\emph{i.e.}, $\times2/3/4$). However, MDSR cannot super-resolve images with non-integer scale factors. Recently, Hu \emph{et al.} \cite{Hu2019Meta} proposed a Meta-SR network to solve the scale-arbitrary upsampling problem. Specifically, they used meta-learning to predict weights of filters for different scale factors. \textcolor{black}{Nevertheless, Meta-SR does not exploit the benefits of scale information during feature learning in the backbone. To make better use of scale information, Fu \emph{et al.} \cite{Fu2021Residual} introduced a residual scale attention network (RSAN), where the scale information is used as a prior knowledge to learn discriminative features for superior performance.}

	\textcolor{black}{Despite Meta-SR and RSAN are able to super-resolve images with non-integer scale factors, they cannot handle asymmetric SR. In many real-world applications like image retargeting, image editing and artworks, asymmetric SR is also highly demanded. However, it is still under-investigated in literature.  In this paper, we develop a plug-in module for existing SR networks to enable SR with both non-integer and asymmetric scale factors.}

	\noindent \textbf{Conditional Convolutions.}
	Different from traditional convolutional layers with static filters, conditional convolutions \cite{Chen2019Dynamic,Yang2019CondConv,Zhang2020DyNet,Tian2020Conditional} parameterize their filters conditioned on the input as linear combinations of several experts. Consequently, the capacity of the network can be efficiently improved without a significant increase in computational cost. In this paper, we extend the idea of conditional convolutions to generate dynamic scale-aware filters to handle the scale-arbitrary SR task. It is demonstrated that  conditional convolutions facilitate our network to adapt to arbitrary scale factors to achieve better SR performance.
	
	\begin{figure*}[t]
		\centering
		\setcounter{figure}{2}
		\includegraphics[width=1\linewidth]{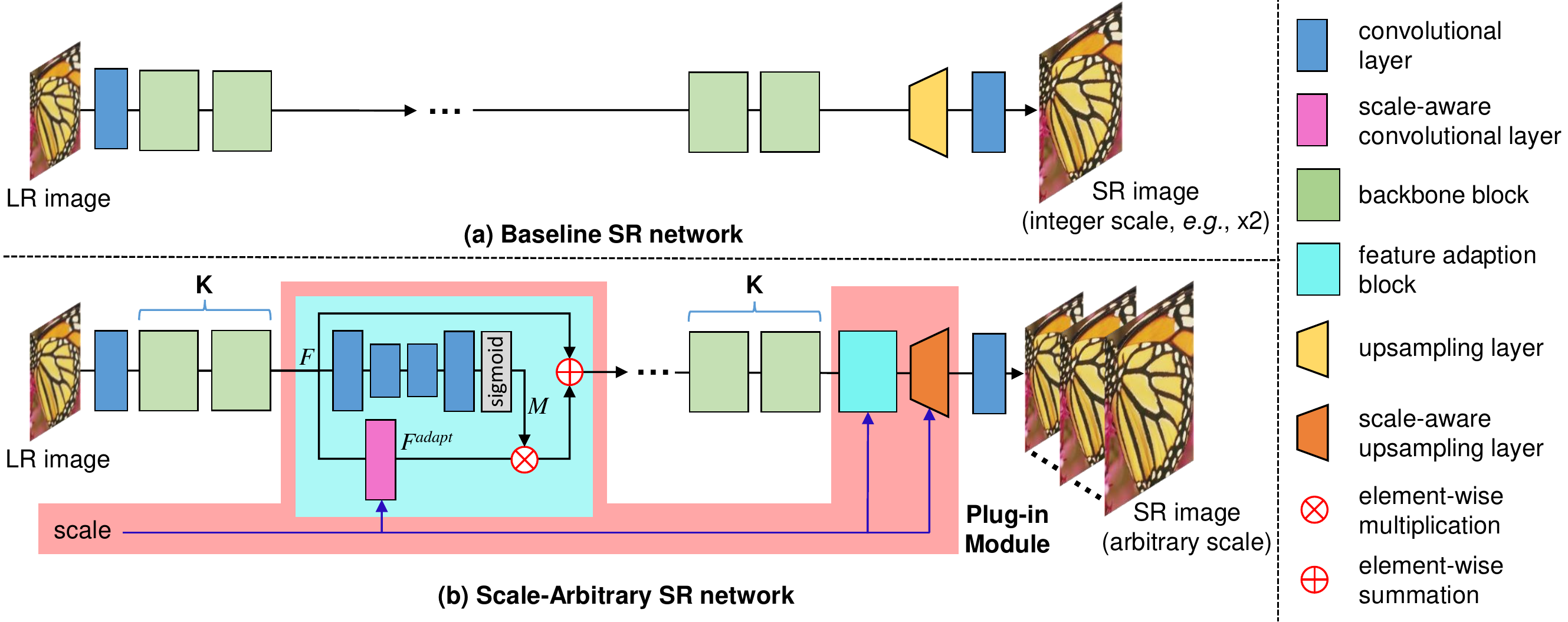}
		\caption{An overview of our plug-in module. The details of our scale-aware convolutional layer and scale-aware upsampling layer are further illustrated in Figs.~\ref{fig5} and \ref{fig6}, respectively.}
		\label{fig2}
		\vspace{-0.2cm}
	\end{figure*}
	\begin{figure}
		\centering
		\setcounter{figure}{1}
		\includegraphics[width=1\linewidth]{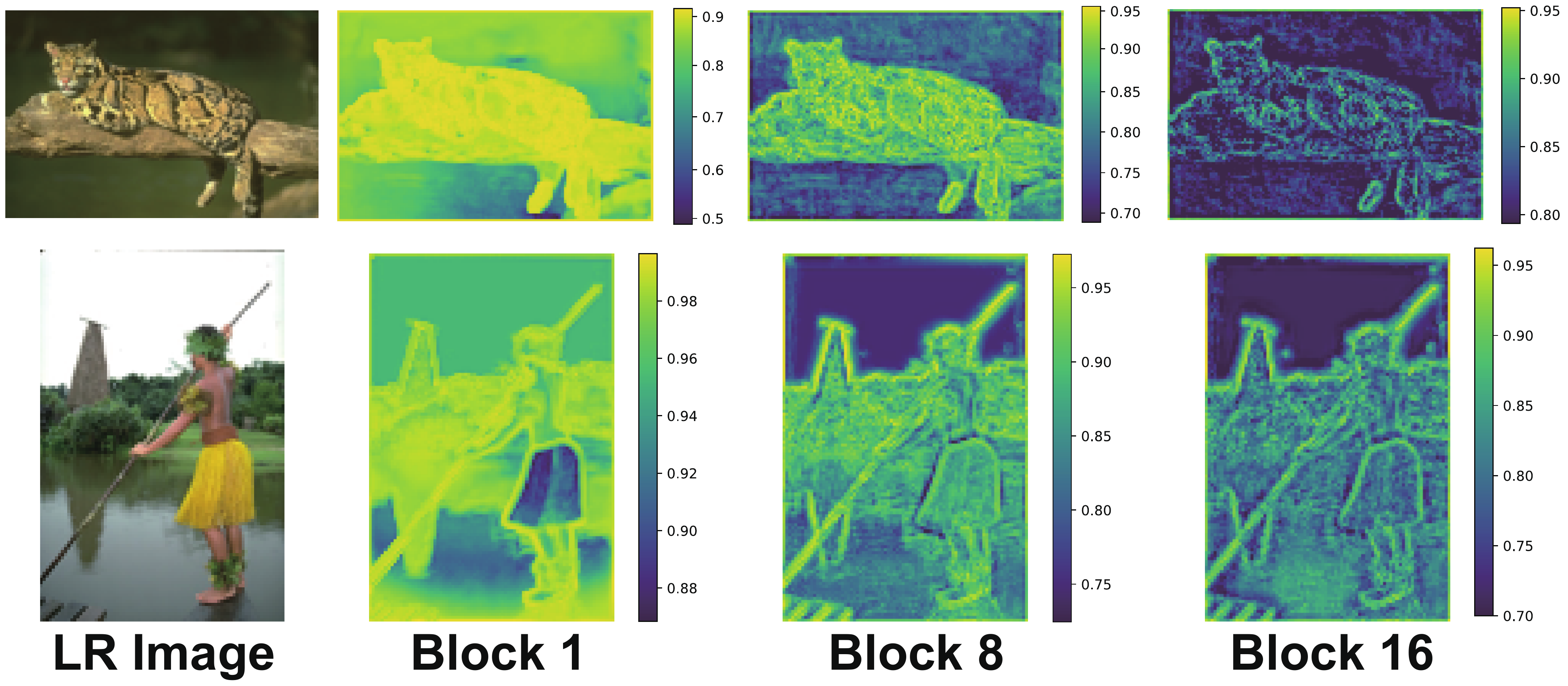}
		\caption{Visualization of feature similarity maps.}
		\label{fig1}
		\vspace{-0.2cm}
	\end{figure}

	\section{Methodology}
	\subsection{Motivation}
	\label{Sec3.1}
	
	Since SR tasks with different scale factors are inter-related \cite{Lim2017Enhanced}, it is non-trivial to learn a scale-arbitrary SR network from scale-specific networks (\emph{e.g.}, $\times2/3/4$). 
	\textcolor{black}{Early attempts \cite{Kim2016Accurate,Dong2016Accelerating,Ahn2018Fast} use shared features in the backbone to handle multiple scale factors without considering the scale information during feature learning. Intuitively, since the degradation is different for various scale factors, scale information can further be used to learn discriminative features to improve SR performance \cite{Fu2021Residual}.}
	In this section, we investigate the relationship between $\times2/3/4$ SR tasks to provide insights for scale-arbitrary SR.
	
	We conduct experiments to compare the feature similarity on specific layers in pre-trained $\times2/3/4$ SR networks. In our experiments, EDSR \cite{Lim2017Enhanced} is selected as the baseline network. 
	First, we downsample an image to $\frac{1}{4}$ size (denoted as $I\!\in\!\mathbb{R}^{H\times{W}}$). Then, we feed $I$ to EDSR networks developed for $\times2/3/4$ SR. 
	Following \cite{Yin2020Disentangled}, features from the last layer in the $i^{\rm th}$ residual block are whitened to remove global component for more precise calculation of pairwise similarity,
	resulting in ${F}^{\times2}_{i},{F}^{\times3}_{i},{F}^{\times4}_{i}\!\in\!\mathbb{R}^{256\times{H}\times{W}}$. 
	\textcolor{black}{Next, for each location $p$, we extract a triplet of feature samples at $p$ to obtain ${f}^{\times2}_{i},{f}^{\times3}_{i},{f}^{\times4}_{i}\in\mathbb{R}^{256}$ and compute the feature similarity among them:} 
	\begin{equation}
	\begin{aligned}
	S_i(p)\!=\!
	\frac{1}{3}\left(\frac{({f}^{\times2}_{i})^{T}{f}^{\times3}_{i}}{\Vert{f}^{\times2}_{i}\Vert\Vert{f}^{\times3}_{i}\Vert}\!+\!
	\frac{({f}^{\times2}_{i})^{T}{f}^{\times4}_{i}}{\Vert{f}^{\times2}_{i}\Vert\Vert{f}^{\times4}_{i}\Vert}\!+\!
	\frac{({f}^{\times3}_{i})^{T}{f}^{\times4}_{i}}{\Vert{f}^{\times3}_{i}\Vert\Vert{f}^{\times4}_{i}\Vert}\right)
	\end{aligned}.
	\end{equation}
	The feature similarity map $S_i$ is visualized in Fig.~\ref{fig1}. For more results, please refer to the supplemental material. 
	From Fig.~\ref{fig1}, we can see that feature similarity varies for different blocks and regions. \textcolor{black}{That is, the sensitivity of features to the change of scale factors is different for various blocks and regions. Consequently, we are motivated to perform pixel-wise feature adaption accordingly.}
	For features within regions of high feature similarities, they can be directly used for SR with arbitrary scale factors. In contrast, features within regions of low feature similarities are adapted to specific scale factors. \textcolor{black}{More analyses are included in the supplemental material.}
	
	
	\subsection{Our Plug-in Module}
	The architecture of our plug-in module is shown in Fig.~\ref{fig2}. Given a baseline network (\emph{e.g.}, EDSR) developed for SR with integer scale factors, we can extend it to a scale-arbitrary SR network using our plug-in module. Specifically, scale-aware feature adaption is performed after every $K$ backbone blocks, as shown in Fig.~\ref{fig2}(b). Following the backbone module, a scale-aware upsampling layer is used for scale-arbitrary upsampling. 
	\begin{figure}[t]
		\centering
		\setcounter{figure}{3}
		\includegraphics[width=1\linewidth]{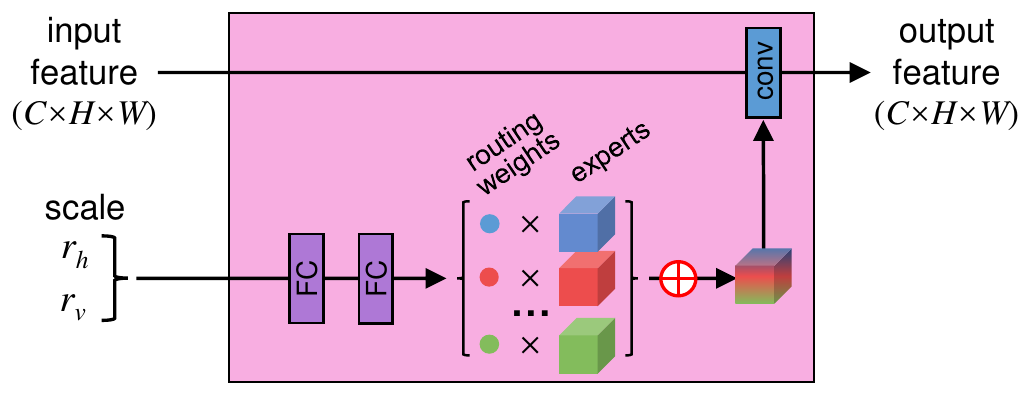}
		\caption{An illustration of our scale-aware convolutional layer.}
		\label{fig5}
		\vspace{-0.1cm}
	\end{figure}
	
	\begin{figure}[t]
		\centering
		\setcounter{figure}{4}
		\includegraphics[width=0.92\linewidth]{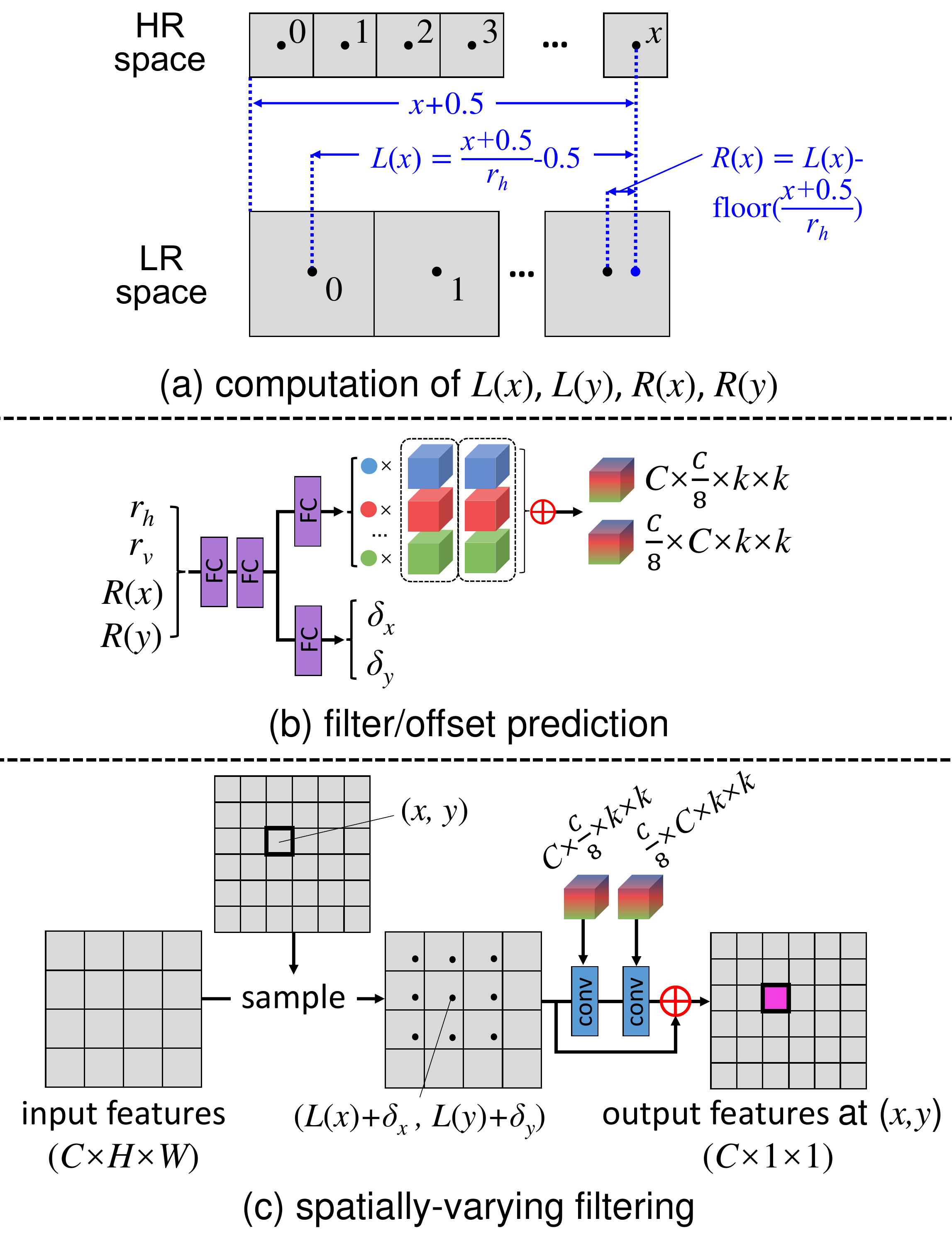}
		\caption{An illustration of our scale-aware upsampling layer.}
		\label{fig6}
		\vspace{-0.1cm}
	\end{figure}
	
	\noindent \textbf{Scale-Aware Feature Adaption.}
	Given a feature map $F$, it is first fed to an hourglass module with four convolutions \textcolor{black}{and a sigmoid layer} to generate a guidance map $M$ with values ranging from 0 to 1, as shown in Fig.~\ref{fig2}(b). Then, $F$ is fed to a scale-aware convolution for feature adaption, resulting in an adapted feature map $F^\emph{adapt}$. \textcolor{black}{Next, the guidance map $M$ is used to fuse $F$ and $F^\emph{adapt}$ as:}
	\begin{equation}
	\label{eq2}
	F^\emph{fuse}=F+F^\emph{adapt}\times{M}.
	\end{equation} 
	\textcolor{black}{Intuitively, in regions with high feature similarities across different scale factors, $F$ can be directly used as $F^\emph{fuse}$. In contrast, in regions with low feature similarities, $F^\emph{adapt}$ is added into $F^\emph{fuse}$ for feature adaption. That is, $M$ serves as a gating mechanism and learns to guide pixel-wise feature adaption. It is demonstrated in Sec.~\ref{Sec4.3} that our network benefits from the guidance maps to produce better results for scale-arbitrary SR.}
	
	The scale-aware convolutional layer within our feature adaption blocks is further illustrated in Fig.~\ref{fig5}. First, 
	the horizontal and vertical scale factors $r_h$ and $r_v$ are fed to a model controller with two fully connected (FC) layers to generate routing weights. Then, these routing weights are used to combine the experts, resulting in a scale-aware filter. \textcolor{black}{Here, experts represent a set of convolutional kernels to be combined based on the scale information.} Finally, the predicted filter is used to process the input feature maps for feature adaption. 
	Different from vanilla convolution with fixed filter, our scale-aware convolution dynamically customizes its filter conditioned on the scale information by combining knowledge from experts. 
	It is demonstrated in Sec.~\ref{Sec4.3} that scale-aware convolutions facilitate our network to adapt to specific scale factors to achieve better performance.

	\noindent \textbf{Scale-Aware Upsampling.}
	Pixel shuffling layer \cite{Shi2016Real} is widely used in SR networks for upsampling with integer scale factors. For $\times{r} (r=2,3,4)$ SR, input features of size $C_{in}\times{H}\times{W}$ are first fed to a convolution to produce features of size $r^2C_{out}\times{H}\times{W}$. Then, the resulting features are shuffled to the size of $C_{out}\times{rH}\times{rW}$. The pixel shuffling layer can be considered as a two-step pipeline, which consists of a sampling step and a spatially-varying filtering step (\emph{i.e.}, $r^2$ convolutions for $r^2$ different sub-positions). Please refer to the supplemental material for more details.
	
	In this paper, we generalize the pixel shuffling layer to a scale-aware upsampling layer, as shown in Fig.~\ref{fig6}. First, each pixel $(x,y)$ in the HR space is projected to the LR space to compute its coordinates ($L(x)$ and $L(y)$) and relative distances ($R(x)$ and $R(y)$):
	\begin{equation}
	L(x)=\frac{x+0.5}{r_h}-0.5,
	\end{equation}
	\begin{equation}
	R(x)=L(x)-{\rm floor}(\frac{x+0.5}{r_h}),
	\end{equation}
	where $L(y)$ and $R(y)$ are calculated similar to $L(x)$ and $R(x)$. Next, $R(x)$, $R(y)$, $r_h$ and $r_v$ are concatenated and fed to two FC layers for feature extraction, as shown in Fig.~\ref{fig6}(b). The resulting features are then passed to filter and offset heads to predict routing weights and offsets ($\delta_x$ and $\delta_y$), respectively. After that, the routing weights are used to combine two groups of experts, resulting in a pair of filters for the bottleneck/expansion layers in Fig.~\ref{fig6}(c). 
	Finally, a $k\!\times\!{k}$ neighborhood centered at $(L(x)+\delta_x,L(y)+\delta_y)$ is sampled \textcolor{black}{using bilinear interpolation} and convolved with the predicted filters to produce the output features at $(x,y)$, as shown in Fig.~\ref{fig6}(c). 
	
	\textcolor{black}{In the implementation, a pair of convolutional kernels ($\mathbb{R}^{C\times{\frac{C}{8}}\times k\times k}$ and $\mathbb{R}^{{\frac{C}{8}}\times C\times k\times k}$) need to be generated and stored for each location in HR space. Since the memory consumption can be very high for $k\!=\!3$ ($\sim$31.6G for a 720P HR image), $k$ is set to 1 in our networks for memory efficiency ($\sim$3.5G).}

	\section{Experiments}
	\subsection{Datasets and Metrics}
	We used the DIV2K dataset \cite{Agustsson2017NTIRE} for network training and five benchmark datasets for evaluation, including Set5 \cite{Bevilacqua2012Low}, Set14 \cite{Zeyde2010Single}, B100 \cite{Martin2001database}, Urban100 \cite{Huang2015Single}, and Manga109 \cite{Matsui2017Sketch}. Peak signal-to-noise ratio (PSNR) and structural similarity index (SSIM) were used as evaluation metrics. Similar to \cite{Hu2019Meta}, we cropped borders for fair comparison. Note that, all metrics were computed in the luminance channel.
	
	\begin{table*}[t]
		\caption{PSNR results achieved by our network with different settings on Set5. }
		\begin{center}
			\begin{threeparttable}
				\renewcommand\arraystretch{1.1}
				\scriptsize
				\setlength{\tabcolsep}{1.5mm}{
					\begin{tabular}{|c|cc|c|ccccc|ccccc|}
						\hline 
						\multirow{2}{*}{Model} &\multicolumn{2}{|c|}{Scale-Aware Feature Adaption}
						& \multirow{2}{*}{\tabincell{c}{Scale-Aware\\Upsampling}}
						& \multirow{2}{*}{{$\times1.7$}}
						& \multirow{2}{*}{{$\times2$}}
						& \multirow{2}{*}{{$\times2.95$}}
						& \multirow{2}{*}{{$\times3$}}
						& \multirow{2}{*}{{$\times3.1$}}
						& \multirow{2}{*}{{$\frac{\times1.3}{\times3.9}$}}
						& \multirow{2}{*}{{$\frac{\times1.9}{\times3.5}$}}
						& \multirow{2}{*}{{$\frac{\times2}{\times3.3}$}}
						& \multirow{2}{*}{{$\frac{\times3.3}{\times1.9}$}}
						& \multirow{2}{*}{{$\frac{\times4}{\times1.8}$}}
						\tabularnewline
						\cline{2-3}
						&{Scale-Aware Conv} & {Guidance Map} &&&&&&&&&&&
						\tabularnewline
						\hline
						\tabincell{c}{EDSR (+Bicubic)*}		&\ding{55}	 &\ding{55}	  &\ding{55}   
						& 39.72 & 38.19 & 34.64 & 34.68 & 34.25 & 34.10 & 34.70 & 34.92 & 35.68 & 34.61
						\tabularnewline
						\hline
						1	& \ding{55}  & \ding{55}  & bicubic    
						& 39.56 & 37.86 & 33.77 & 33.78 & 33.47 & 33.12 & 33.90 & 34.21 & 34.90 & 33.42
						\tabularnewline
						2	& \ding{55}  & \ding{55}  & \ding{51}  
						& 39.76 & 38.13 & 34.70 & 34.68 & 34.40 & 34.32 & 34.91 & 35.02 & 35.85 & 34.67
						\tabularnewline
						3	& \ding{51}  & \ding{55}  & \ding{51}  
						& 39.81 & 38.15 & 34.70 & 34.70 & 34.42 & 34.38 & 34.98 & 35.10 & 35.91 & 34.72
						\tabularnewline
						4	& \ding{51}  & \ding{51}  & \ding{51}  
						& \textbf{39.87} & \textbf{38.19} & \textbf{34.75} & \textbf{34.73} & \textbf{34.48} & \textbf{34.44} & \textbf{35.03} & \textbf{35.16} & \textbf{35.95} & \textbf{34.81}
						\tabularnewline
						\hline
				\end{tabular}}
				\begin{tablenotes}
					\footnotesize
					\item[*] To perform SR with non-integer and asymmetric scale factors (\emph{e.g.}, $\times1.7/\frac{\times1.3}{\times3.9}$ SR) using baseline network, we first super-resolve the LR image for $\times2/\times4$ SR and then downscale the result to the expected resolution using bicubic interpolation following \cite{Hu2019Meta}.
				\end{tablenotes}
			\end{threeparttable}
		\end{center}
		\vspace{-0.6cm}
		\label{tab1}
	\end{table*}
	
	\subsection{Implementation Details}
	Following \cite{Hu2019Meta}, symmetric scale factors varying from 1 to 4 with a stride of 0.1 (\emph{i.e.}, $1.1,1.2,...,3.9,4.0$) were used to generate LR training images. Moreover, asymmetric scale factors with a stride of 0.5 along horizontal and vertical axes (\emph{i.e.}, $\frac{1.5}{2.0},\frac{1.5}{2.5},...,\frac{4.0}{3.0},\frac{4.0}{3.5}$) were also included for LR image generation.
	During training, a pair of horizontal/vertical scale factors was randomly selected from the above ranges for each batch and then 16 LR patches with the size of $50\times50$ were randomly cropped. Meanwhile, their corresponding HR patches were also cropped. Data augmentation was performed through random rotation and random flipping. 
	
	In our experiments, EDSR \cite{Lim2017Enhanced}, RDN \cite{Zhang2018Residual} and RCAN \cite{Zhang2018Image} were used as baseline networks to produce three scale-arbitrary networks, \emph{i.e.}, ArbEDSR, ArbRDN and ArbRCAN. We use 4 experts in the scale-aware convolutions and set $K\!=\!4/2/1$ for ArbEDSR/ArbRDN/ArbRCAN to control the model size. 
	Since the available pre-trained RDN models are implemented in Torch while our networks are implemented in PyTorch \cite{Paszke2019PyTorch}, we re-trained RDN as our baseline network. 
	Pre-trained $\times4$ SR models of EDSR/RDN/RCAN were used to initialize the backbone blocks in ArbEDSR/ArbRDN/ArbRCAN, respectively.
	We used the Adam method \cite{Kingma2015Adam} with $\beta_{1}=0.9$ and $\beta_{2}=0.999$ for optimization. An $L_1$ loss between SR results and HR images was used as the loss function. Following \cite{Zhang2018Residual}, 1000 iterations of back-propagation constitute an epoch. The initial learning rate was set to $1\times10^{-4}$ and reduced to half after every 30 epochs. To maintain training stability, we first trained our networks on integer scale factors ($r=2,3,4$) for 1 epoch and then trained the networks on all scale factors. The training was stopped after 150 epochs.

	\subsection{Ablation Study}
	Ablation experiments were conducted on Set5 to test the effectiveness of our design choices. We used EDSR as the baseline network and introduced 4 variants. All variants were re-trained for 150 epochs. 
	
	\noindent \textbf{Scale-Aware Upsampling.} 
	To enable scale-arbitrary SR, a naive approach is to replace the pixel shuffling layer with an interpolation layer (\emph{e.g.}, bicubic interpolation). 
	To demonstrate the effectiveness of our scale-aware upsampling layer, we introduced two variants. For variant 1, we replaced the pixel shuffling layer in the baseline network with a bicubic upsampling layer. For variant 2, we replaced the pixel shuffling layer with the proposed scale-aware upsampling layer. It can be observed from Table~\ref{tab1} that the PSNR values are relatively low when bicubic upsampling is used. With our scale-aware upsampling layer, the performance is significantly improved (\emph{e.g.}, 39.76/38.13 vs. 39.56/37.86 for $\times1.7/2$ SR). That is because, our scale-aware upsampling layer can learn dynamic filters conditioned on the scale factors while bicubic upsampling uses a fixed filter.
	
	\noindent \textbf{Scale-Aware Feature Adaption.}
	Scale-aware feature adaption is used to adapt features to specific scale factors for better performance. Note that, our scale-aware feature adaption block consists of two key components: scale-aware convolution and guidance map generation. To demonstrate their effectiveness, we first added scale-aware convolutions to variant 2 to produce variant 3. Then, variant 4 is further obtained by adding guidance map generation to variant 3. It can be observed from Table~\ref{tab1} that the performance benefits from both scale-aware convolution and guidance map, with PSNR values being improved from 39.76/38.13/34.91 to 39.87/38.19/35.03 for $\times1.7/2/\frac{1.9}{3.5}$ SR. 
	\textcolor{black}{Without feature adaption, model 2 uses shared features in the backbone for SR with different scale factors. This variant suffers inferior performance since the difference among features learned for various scale factors is not considered.}
	With our scale-aware feature adaption blocks, our network can adapt features in the backbone according to the scale information. Therefore, better performance can be achieved.
	
	We further visualize the guidance maps learned by variant 4 in Fig.~\ref{fig4}. It can be observed that the learned maps are consistent with the feature similarity maps (also shown in Fig.~\ref{fig1}). Specifically, regions with high values in guidance maps are consistent with those of low feature similarities. \textcolor{black}{This demonstrates that our 
		guidance maps can effectively guide the fusion of $F$ and $F^{\emph{adapt}}$ (Eq.~\ref{eq2}) to perform pixel-wise feature adaption accordingly.
	}
	
	\begin{figure}[t]
		\centering
		\includegraphics[width=1\linewidth]{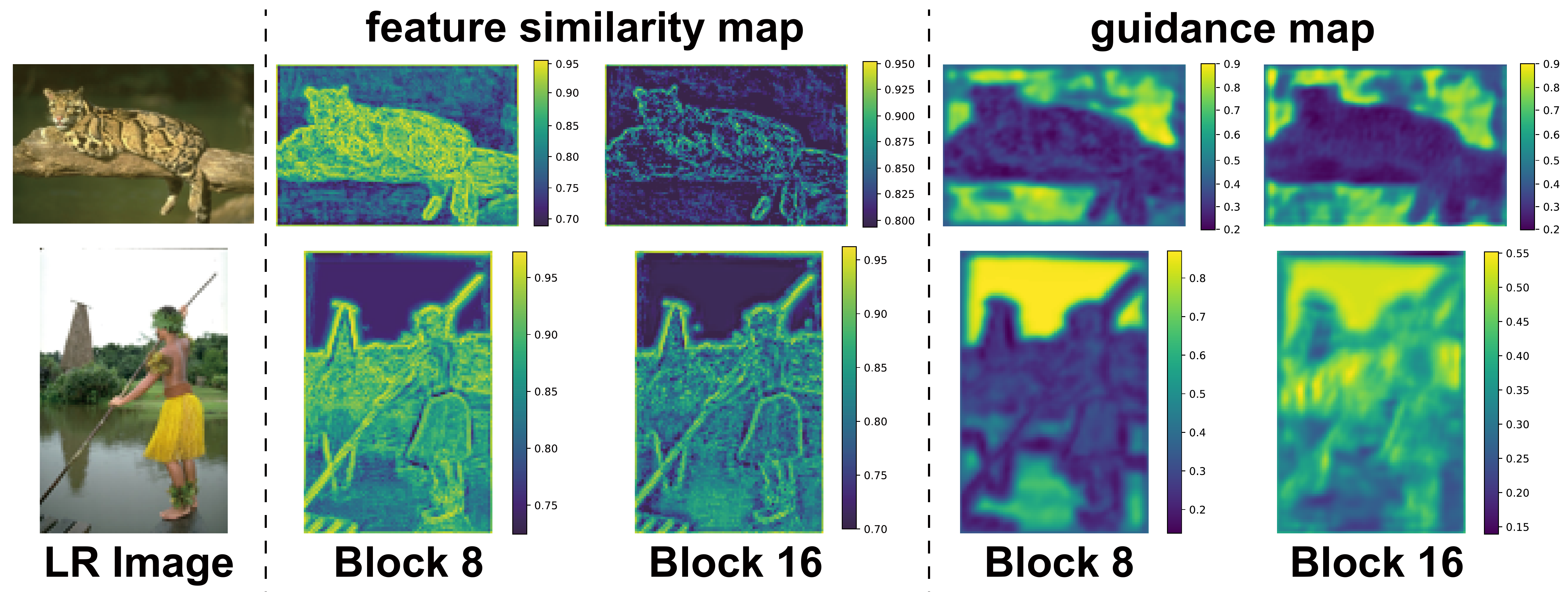}
		\caption{Visualization of guidance maps and their corresponding feature similarity maps.}
		\label{fig4}
	\end{figure}
	
	\begin{table}[t]
		\caption{PSNR results achieved by our network with different number of experts on Set5. The running time is averaged over B100 on $\times4$ SR.}
		\label{tab4}
		\vspace{-0.2cm}
		\begin{center}
			\renewcommand\arraystretch{1.1}
			\scriptsize
			\setlength{\tabcolsep}{0.5mm}{
				\begin{tabular}{|c|c|c|cccc|cccc|}
					\hline 
					\#Experts & Params. & Time 
					& $\times1.7$ & $\times2$ & $\times2.55$ & $\times3.8$ 
					& $\frac{\times1.3}{\times3.9}$ & $\frac{\times2}{\times3.5}$ & $\frac{\times3.3}{\times1.8}$ & $\frac{\times4}{\times1.2}$
					\tabularnewline  
					\hline
					1 & 38.4M & 0.09s & 39.37 & 37.87 & 35.82 & 32.13 & 33.34 & 34.37 & 35.50 & 33.93
					\tabularnewline				
					\hline
					2 & 38.6M & 0.10s & 39.88 & 38.20 & 36.02 & 32.99 & 34.23 & 34.92 & 36.03 & 34.87
					\tabularnewline
					\hline
					4 & 39.2M & 0.10s & 39.87 & 38.19 & 36.02 & 33.00 & 34.44 & 34.96 & 36.07 & 35.18
					\tabularnewline
					\hline
					8 & 40.4M & 0.11s & 39.88 & 38.22 & 36.03 & 32.98 & 34.46 & 34.98 & 36.07 & 35.20
					\tabularnewline					
					\hline
			\end{tabular}}
		\end{center}
		\vspace{-0.5cm}
	\end{table}
	
	\begin{figure*}[t]
		\centering
		\setcounter{figure}{7}
		\includegraphics[width=1\linewidth]{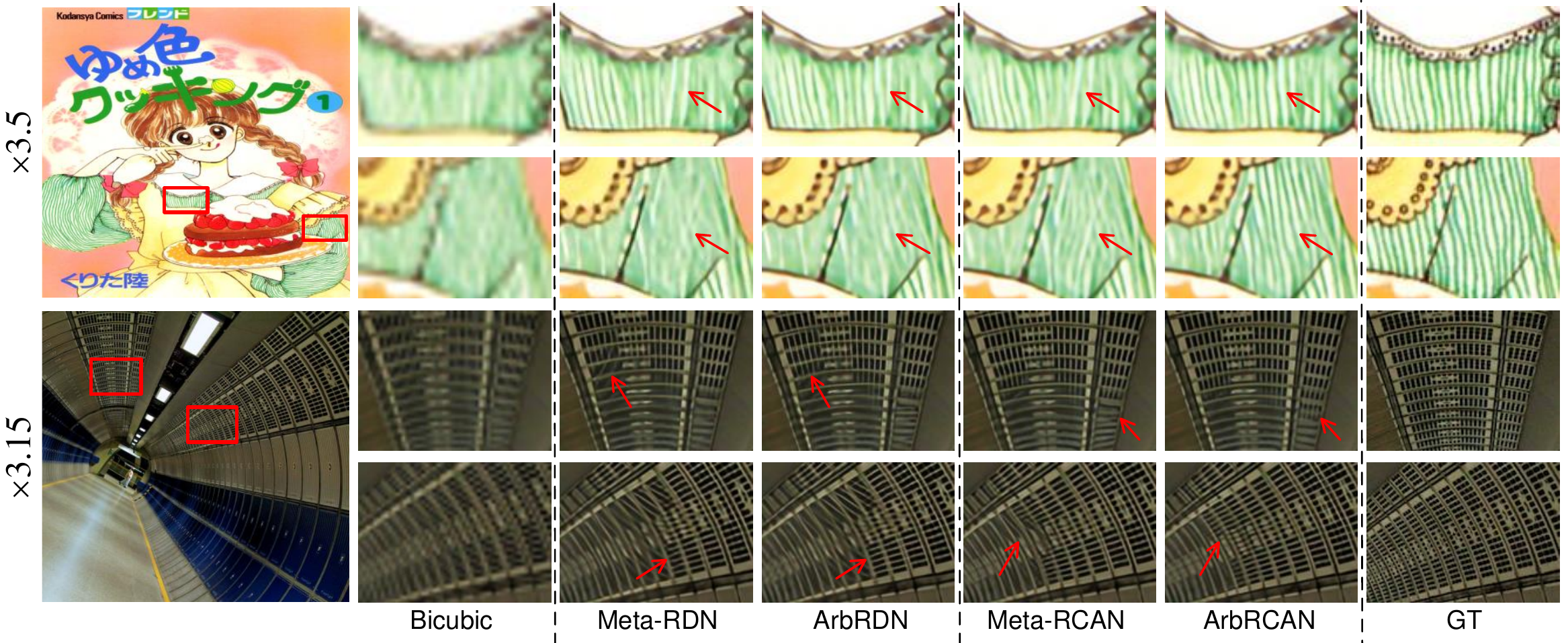}
		\caption{Visual comparison for SR with non-integer scale factors.}
		\label{fig7}
		\vspace{-0.2cm}
	\end{figure*}
	
	\begin{figure}[t]
		\centering
		\setcounter{figure}{6}
		\includegraphics[width=0.8\linewidth]{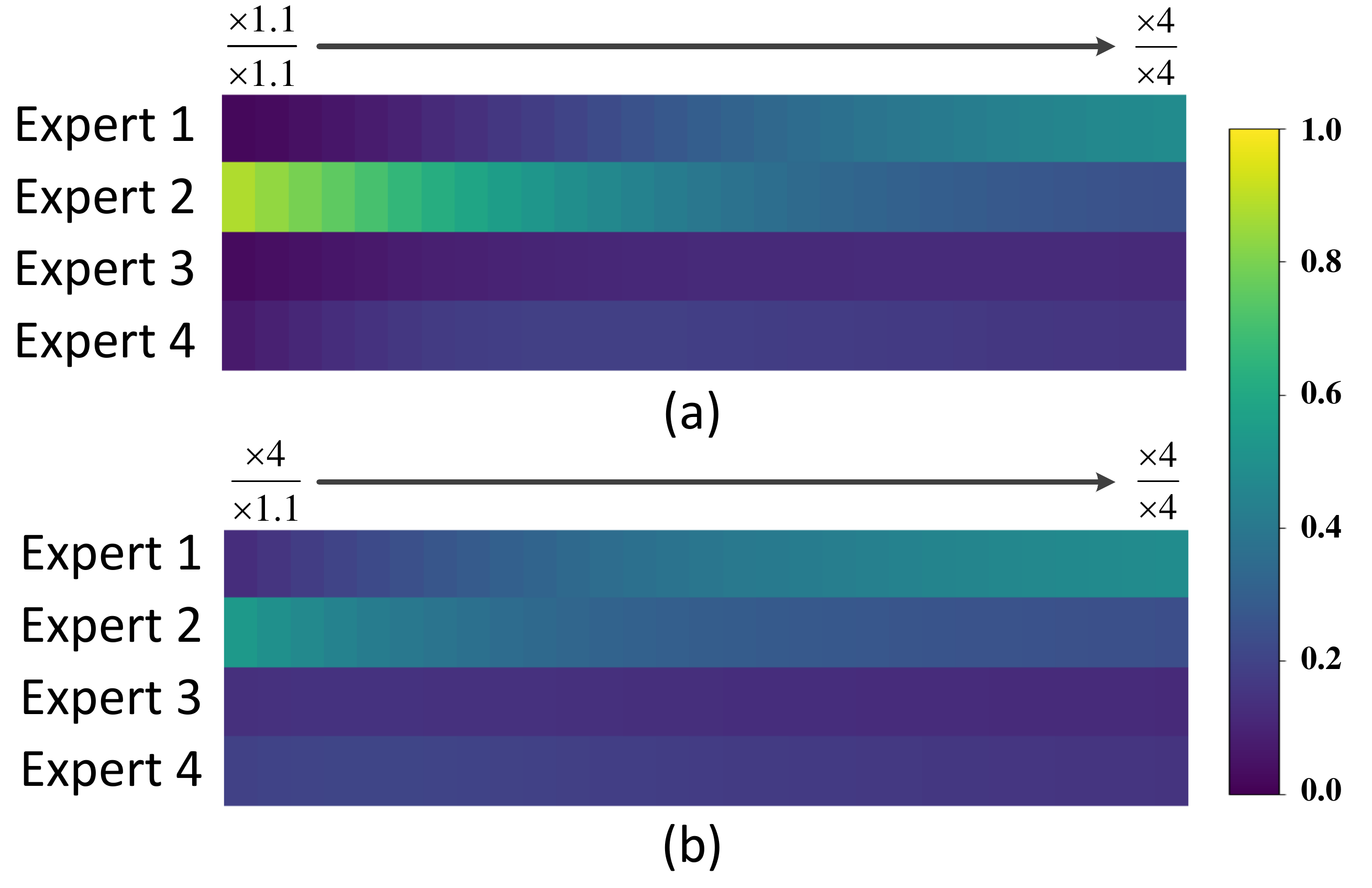}
		\vspace{-0.1cm}
		\caption{Visualization of routing weights in the scale-aware convolution of the first scale-aware feature adaption block. (a) and (b) show the routing weights for symmetric and asymmetric scale factors, respectively.}
		\label{fig10}
		\vspace{-0.5cm}
	\end{figure}


	\noindent \textbf{Number of Experts in Scale-Aware Convolution.}\label{Sec4.3}
	Scale-aware convolution dynamically generates scale-aware filters by combining knowledge from experts. To analyze the effect of the number of experts, we compare the performance of our network with different numbers of experts in Table~\ref{tab4}. With only one expert, our scale-aware convolutions are degraded to vanilla ones with static filters and cannot well handle different scale factors. Therefore, variant 1 suffers a performance drop from 39.88/38.22/36.03 to 39.37/37.87/35.82 for $\times1.7/2/2.55$ SR. 
	As more experts are included in scale-aware convolutions, our network produces comparable results on symmetric scale factors while achieving better performance on asymmetric scale factors. 
	This means that SR with asymmetric scale factors benefits a lot from the increase of experts. 
	Further, we can see that the performance improvement on highly asymmetric scale factors are more significant (\emph{e.g.,} 34.46($\uparrow$0.23)/34.98($\uparrow$0.06) vs. 34.23/34.92 for $\frac{\times1.3}{\times3.9}/\frac{\times2}{\times3.5}$ SR). Since more experts than 4 cannot introduce notable performance improvements, we use 4 experts as our default setting to achieve a better trade-off between performance and model size.

	We further visualize the routing weights in our scale-aware convolution to investigate the knowledge of different experts with respect to various scale factors. As shown in Fig.~\ref{fig10}(a), expert 2 is dominant for small scale factors while expert 1 is gradually activated for large ones. Compared to symmetric scale factors, more experts are activated for asymmetric ones. For example, experts 1, 3 and 4 are assigned with higher weights for $\frac{\times4}{\times1.1}$ than $\frac{\times1.1}{\times1.1}$. This observation is consistent with the results in Table~\ref{tab4} that SR with asymmetric scale factors benefits a lot from more experts.

	\subsection{Results for SR with Symmetric Scale Factors}
	In this section, we compare our ArbEDSR, ArbRDN and ArbRCAN to EDSR \cite{Lim2017Enhanced}, RDN \cite{Zhang2018Residual}, RCAN \cite{Zhang2018Image}, Meta-RDN \cite{Hu2019Meta} and Meta-RCAN \cite{Hu2019Meta} on the SR task with symmetric scale factors (both integer and non-integer scale factors)\footnote{To perform SR with non-integer scale factors (\emph{e.g.}, $\times1.6$) using baseline networks (\emph{e.g.}, RCAN), the LR image is first super-resolved for $\times2$ SR and then bicubicly downscaled to the expected resolution. More analyses are included in the supplemental material.}.  
	Since pre-trained models for Meta-RCAN are unavailable, we used officially released codes for re-training. Note that, we also fine-tuned Meta-RDN and Meta-RCAN on our training set for fair comparison. Comparative results are shown in Table~\ref{tab2} and Fig.~\ref{fig7}.

	\begin{table*}[t]
		\caption{PSNR results achieved on 5 benchmark datasets for symmetric scale factors. Note that, the memory consumption is calculated on an LR image with a size of $100\times100$. The running time is averaged over B100 on $\times2/3/4$ SR. ``+ft'' means that the networks are fine-tuned on our training set.}
		\label{tab2}
		\vspace{-0.6cm}
		\begin{center}
			\renewcommand\arraystretch{1.08}
			\begin{threeparttable}
				\scriptsize
				\setlength{\tabcolsep}{0.85mm}{
					\begin{tabular}{|l|ccc|ccc|ccc|ccc|ccc|ccc|}
						\hline  
						& \multirow{2}{*}{Params.} & \multirow{2}{*}{Memory*} & \multirow{2}{*}{Time*} & \multicolumn{3}{c|}{Set5} & \multicolumn{3}{c|}{Set14} & \multicolumn{3}{c|}{B100}  & \multicolumn{3}{c|}{Urban100} & \multicolumn{3}{c|}{Manga109}
						\tabularnewline
						\cline{5-19}
						& & &
						& $\times2$ & $\times1.6$ & $\times1.55$
						& $\times2$ & $\times1.5$ & $\times1.65$
						& $\times2$ & $\times1.4$ & $\times1.85$
						& $\times2$ & $\times1.9$ & $\times1.95$
						& $\times2$ & $\times1.7$ & $\times1.95$
						\tabularnewline
						\hline
						Bicubic      	
						& - & - & - & 33.66 & 36.10 & 36.24 & 30.24 & 32.87 & 31.83 & 29.56 & 32.95 & 30.11 & 26.88 & 27.25 & 27.05 & 30.80 & 32.91 & 31.12
						\tabularnewline
						\hline
						EDSR-$\times2$ \cite{Lim2017Enhanced}(+Bicubic)  
						& 39.7M & 0.9G & 0.05s & 38.19 & 40.39 & 40.71 & 33.95 & 37.10 & 35.95 & 32.36 & 36.79 & 33.02 & 32.95 & 33.06 & 32.69 & 39.18 & 40.88 & 39.13
						\tabularnewline
						ArbEDSR (Ours)
						& 39.2M & 1.0G & 0.23s & \textbf{38.19} & \textbf{40.64} & \textbf{40.94} & \textbf{34.05} & \textbf{37.51} & \textbf{36.22} & \textbf{32.37} & \textbf{36.92} & \textbf{33.23} & \textbf{33.02} & \textbf{33.61} & \textbf{33.30} & \textbf{39.22} & \textbf{41.20} & \textbf{39.24}
						\tabularnewline
						\hline
						RDN-$\times2$ \cite{Zhang2018Residual}(+Bicubic)  	
						& 21.6M & 0.4G & 0.08s & \textbf{38.24} & 40.51 & 40.53 & 34.01 & 37.24 & 36.10 & 32.34 & 36.83 & 33.15 & 32.89 & 33.05 & 32.79 & 39.18 & 41.06 &39.31
						\tabularnewline
						Meta-RDN \cite{Hu2019Meta}  	 		
						& 21.4M & 1.1G & 0.38s & 38.23 & 40.66 & 40.94 & 34.03 & 37.52 & 36.24 & 32.35 & 36.93 & 33.21 & \textbf{33.03} & 33.60 & 33.26 & 39.31 & 41.33 & 39.60
						\tabularnewline
						Meta-RDN \cite{Hu2019Meta}+ft	 		
						& 21.4M & 1.1G & 0.38s & 38.21 & 40.65 & 40.94 & {34.05} & 37.53 & {36.26} & 32.34 & 36.91 & 33.21 & 33.01 & \textbf{33.61} & \textbf{33.27} & \textbf{39.32} & \textbf{41.35} & \textbf{39.61}
						\tabularnewline
						ArbRDN (Ours)        	
						& 22.6M & 0.6G & 0.18s & 38.23 & \textbf{40.67} & \textbf{40.95} & \textbf{34.07} & \textbf{37.53} & \textbf{36.27} & \textbf{32.37} & \textbf{36.93} & \textbf{33.21} & 33.00 & 33.51 & 33.19 & 39.28 & 41.32 & 39.54
						\tabularnewline
						\hline
						RCAN-$\times2$ \cite{Zhang2018Image}(+Bicubic)	
						& 15.2M & 0.3G & 0.27s & \textbf{38.27} & 40.53 & 40.77 & \textbf{34.12} & 37.23 & 36.08 & \textbf{32.40} & 36.86 & 33.16 & \textbf{33.18} & 33.17 & 32.84 & \textbf{39.42} & 41.15 & 39.39
						\tabularnewline					
						Meta-RCAN \cite{Hu2019Meta}	 		
						& 15.5M & 0.9G & 0.40s & 38.22 & 40.66 & 40.93 & 34.00 & 37.51 & 36.17 & 32.36 & \textbf{36.95} & 33.22  & 33.12 & {33.62} & {33.30} & 39.32 & 41.30 & {39.59}
						\tabularnewline					
						Meta-RCAN \cite{Hu2019Meta}+ft 		
						& 15.5M & 0.9G & 0.40s & 38.21 & 40.63 & 40.93 & 34.03 & 37.50 & 36.20 & 32.35 & 36.95 & 33.22 & 33.10 & \textbf{33.63} & \textbf{33.32} & 39.34 & 41.31 & \textbf{39.61}
						\tabularnewline					
						ArbRCAN (Ours)      	
						& 16.6M & 0.5G & 0.29s & 38.26 & \textbf{40.69} & \textbf{40.97} & 34.09 & \textbf{37.53} & \textbf{36.28} & 32.39 & {36.93} & \textbf{33.23} & 33.14 & 33.55  & 33.25 & 39.37 & \textbf{41.32} & {39.56} 
						\tabularnewline
						\hline
						\hline
						& & &
						& $\times3$ & $\times2.4$ & $\times2.75$
						& $\times3$ & $\times2.8$ & $\times2.95$
						& $\times3$ & $\times2.2$ & $\times2.15$
						& $\times3$ & $\times2.3$ & $\times2.35$
						& $\times3$ & $\times2.7$ & $\times2.55$
						\tabularnewline
						\hline
						Bicubic     	
						& - & - & - & 30.39 & 32.41 & 31.06 & 27.55 & 27.84 & 27.46 & 27.21 & 28.88 & 29.12 & 24.46 & 25.91 & 25.72 & 26.95 & 27.77 & 28.27
						\tabularnewline
						\hline
						EDSR-$\times3$ \cite{Lim2017Enhanced}(+Bicubic)  
						& 42.5M & 1.0G & 0.05s & 34.68 & 36.45 & \textbf{35.35} & 30.53 & 30.90 & 30.49 & 29.27 & 31.38 & \textbf{31.78} & 28.82 & 31.13 & 30.91 & 34.19 & 35.18 & 35.75
						\tabularnewline
						ArbEDSR (Ours)
						& 39.2M & 1.3G & 0.13s & \textbf{34.73} & \textbf{36.54} & {35.34} & \textbf{30.61} & \textbf{31.04} & \textbf{30.56} & \textbf{29.30} & \textbf{31.46} & {31.70} & \textbf{28.90} & \textbf{31.36} & \textbf{31.11} & \textbf{34.28} & \textbf{35.40} & \textbf{36.06}
						\tabularnewline
						\hline
						RDN-$\times3$ \cite{Zhang2018Residual}(+Bicubic)  	
						& 21.7M & 0.4G & 0.08s & 34.71 & 36.46 & 35.27 & 30.57 & 30.88 & 30.53 & 29.26 & 31.30 & 31.65 & 28.80 & 31.25 & 31.07 & 34.13 & 35.41 & 36.00
						\tabularnewline
						Meta-RDN \cite{Hu2019Meta}  	 		
						& 21.4M & 1.9G & 0.32s & \textbf{34.73} & 36.55 & 35.33 & 30.58 & 30.97 & 30.57 & 29.30 & 31.41 & 31.69 & \textbf{28.93} & 31.33 & 31.13 & 34.40 & 35.58 & 36.21
						\tabularnewline
						Meta-RDN \cite{Hu2019Meta}+ft  	 		
						& 21.4M & 1.9G & 0.32s & 34.70 & {36.55} & {35.35} & 30.58 & 30.97 & 30.57 & 29.28 & 31.42 & 31.67 & 28.88 & 31.33 & 31.12 & 34.42 & 35.59 & \textbf{36.22}
						\tabularnewline
						ArbRDN (Ours)        	
						& 22.6M & 0.8G & 0.13s & 30.71 & \textbf{36.55} & \textbf{35.35} & \textbf{30.59} & \textbf{30.98} & \textbf{30.58} & \textbf{29.30} & \textbf{31.45} & \textbf{31.69} & 28.86 & \textbf{31.33} & \textbf{31.14} & \textbf{34.43} & \textbf{35.60} & 36.20
						\tabularnewline
						\hline
						RCAN-$\times3$ \cite{Zhang2018Image}(+Bicubic) 	
						& 15.3M & 0.3G  & 0.27s & 34.76 & 36.51 & 35.31 & 30.62 & 30.90 & 30.53 & 29.31 & 31.31 & 31.68 & \textbf{29.01} & 31.34 & 31.15 & 34.42  & 35.50 & 36.06
						\tabularnewline
						Meta-RCAN \cite{Hu2019Meta} 	 		
						& 15.5M & 1.7G & 0.41s & {34.76} & 36.58 & 35.36 & 30.58 & 31.00 & 30.56 & 29.29 & 31.44 & 31.70 & 28.96 & {31.43} & 31.20 & 34.40  & 35.55 & 36.21
						\tabularnewline
						Meta-RCAN \cite{Hu2019Meta}+ft 	 		
						& 15.5M & 1.7G & 0.41s & 34.72 & 36.59 & 35.38 & 30.58 & 30.99 & 30.56 & 29.28 & 31.46 & 31.70 & 28.93 & 31.44 & 31.22 & 34.44 & 35.60 & 36.24
						\tabularnewline
						ArbRCAN (Ours)       	
						& 16.6M & 0.8G & 0.29s & \textbf{34.76} & \textbf{36.59} & \textbf{35.39} & \textbf{30.64} & \textbf{31.01} & \textbf{30.59} & \textbf{29.32} & \textbf{31.48} & \textbf{31.72} & {28.98} & \textbf{31.48} & \textbf{31.26} & \textbf{34.55} & \textbf{35.64} & \textbf{36.27}
						\tabularnewline
						\hline
						\hline
						& & &
						& $\times4$ & $\times3.1$ & $\times3.25$
						& $\times4$ & $\times3.2$ & $\times3.95$
						& $\times4$ & $\times3.2$ & $\times3.55$
						& $\times4$ & $\times3.7$ & $\times3.85$
						& $\times4$ & $\times3.4$ & $\times3.65$
						\tabularnewline
						\hline
						Bicubic      	
						& - & - & -  & 28.42 & 29.89 & 29.21 & 26.00 & 26.98 & 25.68 & 25.96 & 26.91 & 26.32 & 23.14 & 23.38 & 23.14 & 24.89 & 25.97 & 25.41
						\tabularnewline
						\hline
						EDSR-$\times4$ \cite{Lim2017Enhanced}(+Bicubic)  
						& 42.1M & 1.2G & 0.05s & 32.47 & 34.25 & 33.35 & 28.81 & 29.95 & 28.63 & 27.73 & 28.84 & 28.25 & \textbf{26.65} & 27.06 & 26.69 & 31.04 & 32.51 & 31.79
						\tabularnewline
						ArbEDSR (Ours)
						& 39.2M & 1.7G & 0.10s & \textbf{32.51} & \textbf{34.48} & \textbf{33.92} & \textbf{28.83} & \textbf{30.07} & \textbf{28.72} & \textbf{27.74} & \textbf{28.91} & \textbf{28.30} & 26.62 & \textbf{27.12} & \textbf{26.73} & \textbf{31.26} & \textbf{32.90} & \textbf{32.14}
						\tabularnewline
						\hline
						RDN-$\times4$ \cite{Zhang2018Residual}(+Bicubic)  
						& 21.7M & 0.3G & 0.07s & 32.47 & 34.36 & 33.91 & 28.81 & 30.01 & 28.69 & 27.72 & 28.85 & 28.25 & 26.61 & 27.17 & 26.83 & 31.00 & 32.70 &31.99
						\tabularnewline
						Meta-RDN \cite{Hu2019Meta}  	 		
						& 21.4M & 2.6G & 0.29s & \textbf{32.49} & 34.42 & \textbf{33.93} & 28.86 & 30.06 & \textbf{28.75} & 27.75 & 28.90 & \textbf{28.31} & \textbf{26.70} & \textbf{27.24} & \textbf{26.91} & 31.34 & 33.02 & 32.24
						\tabularnewline
						Meta-RDN \cite{Hu2019Meta}+ft  	 		
						& 21.4M & 2.6G & 0.29s & 32.46 & 34.41 & 33.91 & \textbf{28.86} & {30.06} & 28.74 & \textbf{27.75} & 28.90 & 28.30 & 26.68 & 27.20 &{26.87} & 31.35 & \textbf{33.02} & 32.24
						\tabularnewline
						ArbRDN (Ours)        
						& 22.6M & 1.2G & 0.13s & 32.42 & \textbf{34.43} & 33.92 & 28.82 & \textbf{30.08} & 28.71 & 27.73 & \textbf{28.90} & 28.30 & 26.61 & 27.15 & 26.85 & \textbf{31.35} & 32.99 & \textbf{32.24}
						\tabularnewline
						\hline
						RCAN-$\times4$ \cite{Zhang2018Image}(+Bicubic) 
						& 15.2M & 0.3G & 0.23s & \textbf{32.63} & 34.37 & 33.92 & {28.85} & 30.00 & 28.72 & 27.75 & 28.86 & 28.27 & \textbf{26.75} & 27.20 & 26.89 & 31.20 & 32.76 &32.04
						\tabularnewline					
						Meta-RCAN \cite{Hu2019Meta} 	 
						& 15.5M & 3.1G & 0.39s & 32.56 & 34.46 & 33.98 & 28.85 & 30.08 & 28.73 & 27.75 & 28.86 & 28.30 & 26.71 & \textbf{27.25} & \textbf{26.93} & 31.33 & 33.00 & 32.22
						\tabularnewline	
						Meta-RCAN \cite{Hu2019Meta}+ft 	 
						& 15.5M & 3.1G & 0.39s & 32.55 & 34.44 & 33.99 & 28.85 & 30.08 & 28.73 & 27.75 & 28.88 & 28.30 & 26.71 & 27.24 & 26.93 & 31.35 & 33.02 & 32.23
						\tabularnewline				
						ArbRCAN (Ours)       
						& 16.6M & 1.1G & 0.29s & 32.55 & \textbf{34.50} & \textbf{34.03} & \textbf{28.87} & \textbf{30.08} & \textbf{28.74} & \textbf{27.76} & \textbf{28.93} & \textbf{28.33} & 26.68 & 27.22 & 26.90 & \textbf{31.36} & \textbf{33.12} & \textbf{32.29}
						\tabularnewline
						\hline
				\end{tabular}}
				\begin{tablenotes}
					\footnotesize
					\item[*] Officially released codes for Meta-RDN and Meta-RCAN are used to test the memory consumption and running time. Since generating an input matrix for the weight prediction network (Line 224 of \emph{trainer.py} in the Github repository of Meta-SR) and post-processing (Line 235 of \emph{trainer.py}) are also necessary for a single inference of Meta-SR, these operations are included for a fair comparison of running time.
				\end{tablenotes}	
			\end{threeparttable}
		\end{center}
		\vspace{-0.55cm}
	\end{table*}

	\noindent \textbf{Quantitative Results.}
	It can be observed from Table~\ref{tab2} that our ArbEDSR, ArbRDN and ArbRCAN achieve comparable performance to their corresponding baseline networks on integer scale factors. For SR with non-integer scale factors, our networks significantly outperform their baseline networks. For example, our ArbEDSR is on par with EDSR for $\times2$ SR on Set5 (38.19 vs. 38.19) while producing much better results for $\times1.6/1.55$ SR (40.64/40.94 vs. 40.39/40.71). 
	
	Compared to Meta-RDN and Meta-RCAN, our ArbRDN and ArbRCAN achieves comparable or better performance for most scale factors. For example, our ArbRCAN produces notable performance improvements for $\times3.4/3.65$ SR on Manga109 (33.12/32.29 vs. 33.00/32.22). 
	Moreover, our ArbRDN and ArbRCAN achieve much better efficiency than Meta-RDN and Meta-RCAN, respectively. Compared to RCAN, Meta-RCAN has a comparable model size with larger memory consumption and longer running time. In contrast, our ArbRCAN takes shorter running time (0.29s vs. 0.39s) and much less memory consumption (1.1G vs. 3.1G). This clearly demonstrates the high efficiency of our plug-in module. 

	\noindent \textbf{Qualitative Results.}
	Figure~\ref{fig7} compares the visual results achieved on two images of the Manga109 and Urban100 datasets. From the zoom-in regions, we can see that our ArbRCAN produces results with better perceptual quality and fewer artifacts. For the second test image, Meta-RDN and Meta-RCAN cannot faithfully recover the stripes and suffer distorted artifacts. In contrast, our ArbRCAN produces clearer and finer details.

	\begin{table*}[t]
		\caption{PSNR results achieved for asymmetric scale factors. Note that, the memory consumption is calculated on an LR image with a size of $100\times100$ for $\frac{\times2}{\times4}$ SR. The running time is averaged over B100 on $\frac{\times2}{\times4}$ SR.}
		\vspace{-0.3cm}
		\label{tab3}
		\begin{center}
			\renewcommand\arraystretch{1.1}
			\scriptsize
			\setlength{\tabcolsep}{0.7mm}{
				\begin{tabular}{|l|ccc|ccc|ccc|ccc|ccc|ccc|}
					\hline 
					& \multirow{2}{*}{Params.} & \multirow{2}{*}{Memory} & \multirow{2}{*}{Time}
					& \multicolumn{3}{c|}{Set5} & \multicolumn{3}{c|}{Set14} & \multicolumn{3}{c|}{B100}   & \multicolumn{3}{c|}{Urban100} & \multicolumn{3}{c|}{Manga109}
					\tabularnewline 
					& & & 
					& $\frac{\times1.5}{\times4}$ & $\frac{\times1.5}{\times3.5}$ & $\frac{\times1.6}{\times3.05}$ 
					& $\frac{\times4}{\times2}$ & $\frac{\times3.5}{\times2}$ & $\frac{\times3.5}{\times1.75}$ 
					& $\frac{\times4}{\times1.4}$ & $\frac{\times1.5}{\times3}$ & $\frac{\times3.5}{\times1.45}$ 
					& $\frac{\times1.6}{\times3}$ & $\frac{\times1.6}{\times3.8}$ & $\frac{\times3.55}{\times1.55}$ 
					& $\frac{\times2.5}{\times2}$ & $\frac{\times2.8}{\times3.5}$ & $\frac{\times3.35}{\times2.7}$ 
					\tabularnewline
					\hline
					Bicubic 		  
					& - & - & - & 30.01 & 30.83 & 31.40 & 27.25 & 27.88 & 27.27 & 27.45 & 28.86 & 27.94 & 25.93 & 24.92 & 25.19 & 29.61 & 26.47 & 26.86
					\tabularnewline
					\hline
					EDSR \cite{Lim2017Enhanced}+Bicubic
					& 42.1M & 0.7G & 0.04s & 33.95 & 34.89 & 35.59 & 30.29 & 30.91 & 31.36 & 29.33 & 31.24 & 29.96 & 30.61 & 28.77 & 29.23 & 37.08 & 32.99 & 33.46
					\tabularnewline
					ArbEDSR (Ours)
					& 39.2M & 0.9G & 0.14s & \textbf{34.32} & \textbf{35.33} & \textbf{36.02} & \textbf{30.51} & \textbf{31.15} & \textbf{31.46} & \textbf{29.52} & \textbf{31.38} & \textbf{30.20} & \textbf{31.06} & \textbf{29.32} & \textbf{29.98} & \textbf{37.70} & \textbf{33.54} & \textbf{34.16}
					\tabularnewline
					\hline
					RDN \cite{Zhang2018Residual}+Bicubic 	  
					& 21.7M & 0.4G & 0.08s & 34.12 & 35.04 & 35.63 & 30.32 & 31.02 & 31.16 & 29.34 & 31.29 & 29.98 & 30.68 & 28.75 & 29.30 & 37.43 & 33.27 & 33.77
					\tabularnewline
					Meta-RDN \cite{Hu2019Meta}+Bicubic  
					& 21.4M & 3.1G & 0.49s & 34.19 & 35.17 & 35.79 & 30.39 & 31.06 & 31.36 & 29.43 & 31.28 & 30.09 & 30.77 & 29.04 & 29.63 & 37.74 & 33.61 & 34.22
					\tabularnewline
					Meta-RDN \cite{Hu2019Meta}+Bicubic+ft  
					& 21.4M & 3.1G & 0.49s & 34.22 & 35.19 & 35.80 & 30.42 & 31.06 & 31.35 & 29.47 & 31.30 & 30.12 & 30.85 & 29.11 & 29.70 & 37.80 & 33.64 & 34.26
					\tabularnewline
					ArbRDN (Ours)		      
					& 22.6M & 0.7G & 0.13s & \textbf{34.31} & \textbf{35.26} & \textbf{35.98} & \textbf{30.47} & \textbf{31.12} & \textbf{31.42} & \textbf{29.52} & \textbf{31.36} & \textbf{31.19} & \textbf{31.02} & \textbf{29.23} & \textbf{29.91} & \textbf{37.88} & \textbf{33.74} & \textbf{34.36}
					\tabularnewline
					\hline
					RCAN \cite{Zhang2018Image}+Bicubic 	  
					& 15.2M & 0.4G & 0.27s & 34.14 & 35.05 & 35.67 & 30.35 & 31.02 & 31.21 & 29.35 & 31.30 & 29.98 & 30.72 & 28.81 & 29.34 & 37.48 & 33.31 & 33.82
					\tabularnewline					
					Meta-RCAN \cite{Hu2019Meta}+Bicubic 
					& 15.5M & 2.8G & 0.61s & 34.20 & 35.17 & 35.81 & 30.40 & 31.05 & 31.33 & 29.43 & 31.26 & 30.09 & 30.73 & 29.03 & 29.67 & 37.74 & 33.61 & 34.23
					\tabularnewline					
					Meta-RCAN \cite{Hu2019Meta}+Bicubic+ft 
					& 15.5M & 2.8G & 0.61s & 34.26 & 35.24 & 35.86 & 30.46 & 31.10 & 31.40 & 29.47 & 31.30 & 30.14 & 30.86 & 29.14 & 29.75 & 37.80 & 33.67 & 34.28
					\tabularnewline									
					ArbRCAN (Ours)	  
					& 16.6M & 0.7G & 0.29s & \textbf{34.37} & \textbf{35.40} & \textbf{36.05} & \textbf{30.55} & \textbf{31.27} & \textbf{31.54} & \textbf{29.54} & \textbf{31.40} & \textbf{30.22} & \textbf{31.13} & \textbf{29.36} & \textbf{30.04} & \textbf{37.93} & \textbf{33.81} & \textbf{34.41}
					\tabularnewline
					\hline
			\end{tabular}}
		\end{center}
		\vspace{-0.65cm}
	\end{table*}

	\begin{figure*}
		\centering
		\setcounter{figure}{8}
		\includegraphics[width=1\linewidth]{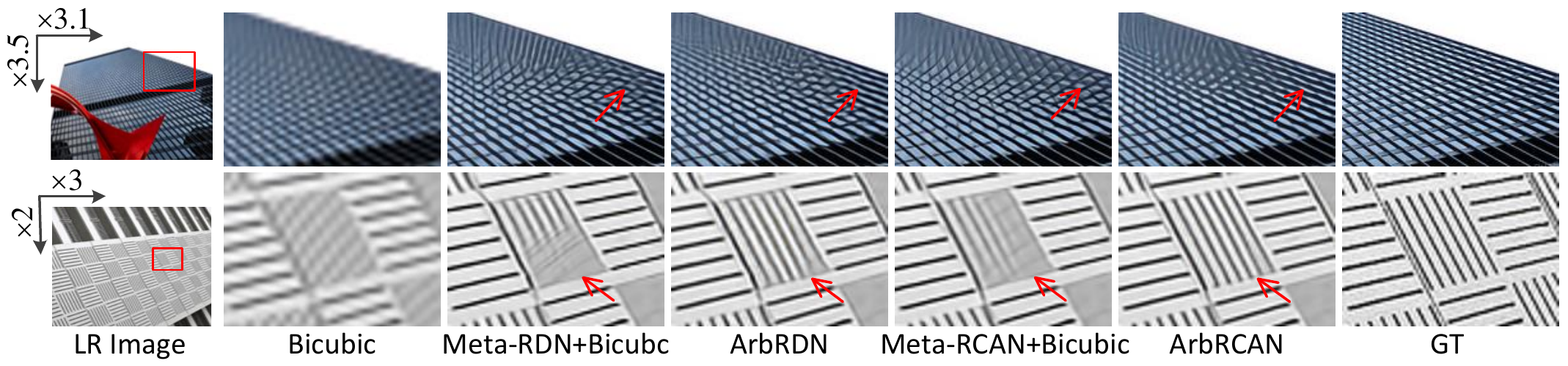}
		\caption{Visual comparison for SR with asymmetric scale factors on Urban100.}
		\label{fig9}
		\vspace{-0.4cm}
	\end{figure*}

	\subsection{Results for SR with Asymmetric Scale Factors}
	In this section, we test our ArbEDSR, ArbRDN and ArbRCAN on the SR task with asymmetric scale factors\footnote{To perform SR with asymmetric scale factors (\emph{e.g.}, $\frac{\times2.5}{\times3.5}$) using baseline networks (\emph{e.g.}, RCAN) and Meta-SR (\emph{e.g.}, Meta-RCAN), the LR image is first super-resolved for $\times4$ and $\times3.5$ SR, respectively. Then, the resultant images are resized to the expected resolution using bicubic interpolation. More analyses  are included in the supplemental material.}. 
	Comparative results are presented in Table~\ref{tab3}, while visual comparison is provided in Fig.~\ref{fig9}.

	\noindent \textbf{Quantitative Results.}
	It can be observed from Table~\ref{tab3} that baseline networks (\emph{e.g.}, RCAN) have limited performance on asymmetric scale factors since their filters are fixed. Meta-RCAN uses meta-learning to generate filters for different scale factors to produce better results, with PSNR values being improved from 37.48/33.31/33.82 to 37.74/33.61/34.23 on Manga109. Moreover, fine-tuning Meta-RCAN on our training set further introduces marginal improvements (37.80/33.67/34.28 vs. 37.74/33.61/34.23). However, the performance is still inferior to our ArbRCAN even after fine-tuning. Using scale-aware convolutions to dynamically customize filters conditioned on the scale information, our ArbRCAN can adapt to the input scale factor to achieve better performance (37.93/33.81/34.41 vs. 37.80/33.67/34.28). 
	
	In addition to higher PSNR results, our ArbRCAN also has much smaller memory cost (0.7G vs. 2.8G) and shorter running time (0.29s vs. 0.61s) as compared to Meta-RCAN. Since Meta-RCAN needs to super-resolve an LR image to a size larger than the expected one before bicubic downscaling to perform asymmetric SR, redundant computational and memory cost is involved. In contrast, our ArbRCAN can directly super-resolve the LR image to the expected size with better efficiency.

	\noindent \textbf{Qualitative Results.}
	Figure~\ref{fig9} illustrates the visual results achieved on two images of the Urban100 dataset. It can be observed from the zoom-in regions that our ArbRCAN produces better results for different asymmetric scale factors. Specifically,  we can see from the second row that, our ArbRDN and ArbRCAN faithfully recover the stripes while other methods suffer blurring artifacts. This further demonstrates the superior performance of our networks on asymmetric scale factors.

	\subsection{Results for SR in the Wild}
\begin{figure}[t]
		\centering
		\includegraphics[width=1.0\linewidth]{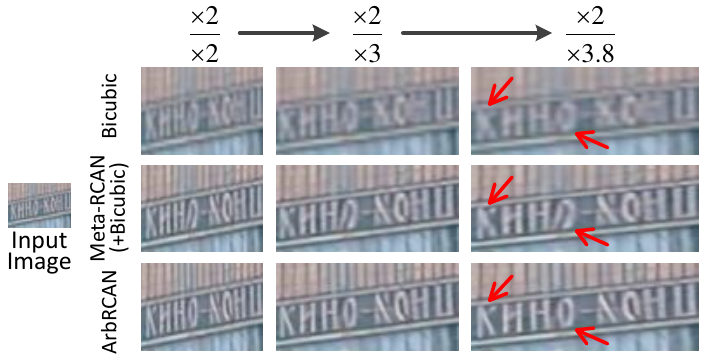}
		\caption{Visual comparison on a real-world image.}
		\label{fig11}
		\vspace{-0.2cm}
	\end{figure}

	In many real-world applications, continuous magnification of an image to an arbitrary size is favored by customers such that they can stretch an image to any resolution they like. However, all existing SR network including RDN, RCAN and Meta-RCAN rely on post-processing (\emph{i.e.}, bicubic interpolation) to achieve scale-arbitrary SR, which is difficult to obtain optimal performance. In contrast, our network provides an end-to-end framework to produce better results. In this section, ArbRCAN is used to super-resolve a real-world image (an HR image in B200 \cite{Martin2001database}) to different resolutions for visual comparison, as shown in Fig.~\ref{fig11}. When the input image is continuously magnified, the text becomes easier for recognition and our ArbRCAN consistently produces better perceptual quality than other methods.
	
	\section{Conclusions}
	In this paper, we proposed a  plug-in module to enable existing image SR networks for scale-arbitrary SR with a single model. 
	Experimental results show that baseline networks equipped with our module can produce promising results on SR tasks with non-integer and asymmetric scale factors, while maintaining state-of-the-art performance on integer scale factors. Moreover, our module can be easily adapted to scale-specific networks with small additional computational and memory cost. 
	
	{\small
		\bibliographystyle{unsrt}
		\bibliography{super-resolution,other-CV-fields,neural-network}
	}

\end{document}